\documentclass{article} 
\usepackage{iclr2025_conference_for_arxiv,times}

\usepackage[hyphens]{url} 
\usepackage{graphicx} 
\usepackage{natbib}  
\usepackage{caption} 

\usepackage{algorithm}
\usepackage{algorithmic}
\usepackage{amsthm}

\usepackage{booktabs}
\usepackage{subcaption}
\usepackage{amsmath}  
\usepackage{hyperref}  
\usepackage{cleveref}  

\newtheorem{theorem}{Theorem}[section]

\newtheorem{corollary}[theorem]{Corollary}
\theoremstyle{definition}

\theoremstyle{remark}
\newtheorem{remark}[theorem]{Remark}
\usepackage[skip=0.333\baselineskip]{caption}
\usepackage{bbold}
\usepackage{amssymb}
\usepackage{wrapfig}

\crefname{section}{section}{sections}
\Crefname{section}{Section}{Sections}

\crefname{subsection}{subsection}{subsections}
\Crefname{subsection}{Subsection}{Subsections}

\crefname{subsubsection}{subsubsection}{subsubsections}
\Crefname{subsubsection}{Subsubsection}{Subsubsections}

\crefname{equation}{equation}{equations}
\Crefname{equation}{Equation}{Equations}

\crefname{figure}{figure}{figures}
\Crefname{figure}{Figure}{Figures}

\crefname{table}{table}{tables}
\Crefname{table}{Table}{Tables}

\crefname{theorem}{theorem}{theorems}
\Crefname{theorem}{Theorem}{Theorems}

\crefname{lemma}{lemma}{lemmas}
\Crefname{lemma}{Lemma}{Lemmas}

\crefname{corollary}{corollary}{corollaries}
\Crefname{corollary}{Corollary}{Corollaries}

\crefname{definition}{definition}{definitions}
\Crefname{definition}{Definition}{Definitions}

\crefname{proposition}{proposition}{propositions}
\Crefname{proposition}{Proposition}{Propositions}

\crefname{remark}{remark}{remarks}
\Crefname{remark}{Remark}{Remarks}

\crefname{example}{example}{examples}
\Crefname{example}{Example}{Examples}

\crefname{algorithm}{algorithm}{algorithms}
\Crefname{algorithm}{Algorithm}{Algorithms}

\usepackage{xspace}
\usepackage{microtype}

\DeclareMathOperator*{\argmin}{arg\,min}
\newcommand{\proj}{\text{proj}}
\newcommand{\prox}{\text{prox}}

\newcommand{\sX}{\mathcal{X}}

\newcommand{\bbE}{\mathbb{E}}
\newcommand{\bbFed}{\text{FedMLS}\xspace}
\newcommand{\bbMLS}{\text{MLS}}
\newcommand{\bbT}{\text{Federated} \text{Learning} \text{with} \text{Multiple} \text{Local} \text{Steps}~}
\newcommand{\norm}[1]{\left\|#1\right\|}

\title{Provable Reduction in Communication Rounds for Non-Smooth Convex Federated Learning}

\author{Karlo Palenzuela\\
Department of Computing Science\\
Umeå University, Umeå, Sweden\\
\texttt{karlo.palenzuela@umu.se} \\
\And
Ali Dadras \& Alp Yurtsever\thanks{Authors made an equal contribution to this work.}\\
Department of Mathematics and Mathematical Statistics \\
Umeå University, Umeå, Sweden\\
\texttt{\{ali.dadras,alp.yurtsever\}@umu.se} \\
\AND
Tommy Löfstedt\footnotemark[1]\\
Department of Computing Science\\
Umeå University, Umeå, Sweden\\
\texttt{tommy.lofstedt@umu.se}
}

\iclrfinalcopy 

\usepackage{xcolor}
\definecolor{brightmaroon}{rgb}{0.76, 0.13, 0.28}
\definecolor{goldenrod}{rgb}{0.85, 0.65, 0.13}
\definecolor{amethyst}{rgb}{0.6, 0.4, 0.8}
\definecolor{applegreen}{rgb}{0.55, 0.71, 0.0}
\definecolor{brightturquoise}{rgb}{0.03, 0.91, 0.87}
\definecolor{ao}{rgb}{0.0, 0.0, 1.0}
\definecolor{alizarin}{rgb}{0.82, 0.1, 0.26}

\begin{document}

\maketitle

\begin{abstract}

Multiple local steps are key to communication-efficient federated learning. However, theoretical guarantees for such algorithms, without data heterogeneity-bounding assumptions, have been lacking in general non-smooth convex problems. Leveraging projection-efficient optimization methods, we propose FedMLS, a federated learning algorithm with provable improvements from multiple local steps. FedMLS attains an $\epsilon$-suboptimal solution in $\mathcal{O}(1/\epsilon)$ communication rounds, requiring a total of $\mathcal{O}(1/\epsilon^2)$ stochastic subgradient oracle calls. 
\end{abstract}

\section{Introduction}

A typical FL algorithm consists of two main phases: local training and aggregation.  Clients update their model parameters using their own data in the local training phase; and these models are collected, combined, and distributed back to the clients in the aggregation phase.  The communication cost is often the primary bottleneck in real-world FL applications, particularly when the number of clients is large, the model has many parameters, or network connections are poor~\citep{konevcny2016federated}. 
A common strategy to address this issue is to allow clients to perform multiple local training steps before aggregation; we refer to this approach as Multiple Local Steps~(\bbMLS{}).

\bbMLS{} can substantially reduce communication costs by lowering the frequency of model transmissions, and has been widely adopted in nearly all FL algorithms.  However, despite its empirical success, theoretically characterizing the improvements in communication cost from \bbMLS{} remains challenging.  Most analyses rely on restrictive and often unverifiable assumptions, such as bounded gradient dissimilarity among clients, which can even conflict with standard conditions like smoothness or strong convexity~\citep{Khaled2019TighterTF}.

Scaffold \citep{Karimireddy2019SCAFFOLDSC} and Scaffnew \citep{Mishchenko2022ProxSkipYL} stand out as notable exceptions in this context. Without relying on any data heterogeneity-bounding assumptions, Scaffold achieves a provable reduction in the number of communication rounds for smooth convex minimization in the stochastic setting, while Scaffnew extends these results to smooth and strongly convex minimization in both deterministic and stochastic settings (see Section 2 for details).

We explore the following natural question in this work: 
\begin{center}
    \emph{Can multiple local steps provably reduce communication rounds in the non-smooth convex setting?}
\end{center}
We answer this question affirmatively by leveraging projection-efficient methods~\citep{Thekumparampil2020ProjectionES} and adapting them to the FL context, leading to our proposed algorithm, \bbFed{}.

Specifically, for convex and Lipschitz continuous functions (but not necessarily Lipschitz smooth), \bbFed{} achieves an $\epsilon$-suboptimal solution in $\mathcal{O}(\epsilon^{-1})$ communication rounds, with $\mathcal{O}(\epsilon^{-1})$ local training steps per round, resulting in a total of $\mathcal{O}(\epsilon^{-2})$ local training steps; see \Cref{sec:mainresult}. The same complexity also holds in expectation for the stochastic setting with unbiased stochastic subgradient estimators of bounded variance. 

The rest of the paper is organized as follows: 
We provide a brief discussion of related work in \Cref{sec:related-works}. 
In \Cref{sec:mainresult}, we develop \bbFed{} and present its theoretical guarantees.  
In \Cref{sec:numerical_exp}, we report numerical experiments on a binary classification task with a Support Vector Machine (SVM), validating the theoretical findings.
Finally, we conclude with a discussion of limitations and future directions.

\section{Related Works} \label{sec:related-works}

Due to space constraints, this section focuses specifically on methods that reduce the frequency of client-server communication. 
Beyond this, researchers have explored various techniques to minimize the communication footprint of FL algorithms, such as sparse or low-rank approximations, quantization, and asynchronous updates.
For a comprehensive review, we refer to \citep{shahid2021communication,zhao2023towards} and references therein. 

\noindent%
\textbf{\bbMLS{} with data heterogeneity-bounding assumptions.~} 
\bbMLS{} has been a key component of FL applications \citep{McMahan2016CommunicationEfficientLO,konevcny2016federated,kairouz2021advances}, however, its analysis with heterogeneous data distributions presents new challenges and remains an active area of research. 
As \citet{zhao2018federated} identified, \bbMLS{} with classical methods like Federated Averaging (FedAvg, aka.~Local SGD) may deviate from the true solution when the data are heterogeneous across clients, a phenomenon known as \textit{client drift}. 
Early works to mitigate this issue introduced various regularity assumptions to bound data heterogeneity, such as bounded gradient or gradient dissimilarity \citep{Stich2018LocalSC,wang2019adaptive,Gorbunov2020LocalSU}, bounded Hessian dissimilarity \citep{Karimireddy2019SCAFFOLDSC}, bounds on the variation in the local gradients at the optimum \citep{Khaled2019FirstAO,woodworth2020minibatch}, or the assumption that clients’ optima are close \citep{Haddadpour2019OnTC,li2020federated}, just to name a few. 
These assumptions are often impractical to verify in FL, particularly since the data are kept locally and are often confidential.

\noindent%
\textbf{\bbMLS{} without data heterogeneity-bounding assumptions.~} 
\citet{Karimireddy2019SCAFFOLDSC} proposed Scaffold, which introduces a control variate that acts as a variance reduction term to mitigate client drift without relying on heterogeneity‐bounding assumptions. 
Scaffold can achieve an $\epsilon$-suboptimal solution in $\smash{\tilde{\mathcal{O}}(\epsilon^{-1})}$ communication rounds for smooth and convex functions, and in $\smash{\tilde{\mathcal{O}}(\kappa)}$ rounds for smooth and strongly convex (where $\tilde{\mathcal{O}}$ hides logarithmic factors). 
Although this improves the communication complexity in the stochastic setting, it does not outperform the complexity of full-batch (\textit{i.e.}, deterministic) updates without \bbMLS{}. 
\citet{Mishchenko2022ProxSkipYL} established the first theoretical improvements in this context. 
They developed a randomized proximal gradient method that skips the prox step with probability $p$ in each iteration. 
When applied to FL, this approach yields an algorithm (Scaffnew) that skips communication steps. 
For smooth and strongly convex functions, they reduced the communication complexity from $\smash{\tilde{\mathcal{O}}(\kappa)}$ to $\smash{\tilde{\mathcal{O}}(\sqrt{\kappa})}$ in the deterministic setting, where $\kappa$ denotes the condition number of the objective, and from $\smash{\mathcal{O}(\epsilon^{-1})}$ to $\smash{\tilde{\mathcal{O}}(\epsilon^{-1/2})}$ in the stochastic setting.
More recently, \citet{Hu2023TighterAF} provided a tighter analysis for Scaffnew, further improving the communication complexity in the stochastic setting to $\tilde{\mathcal{O}}(\sqrt{\kappa})$. 
Similarly, \citet{Maranjyan2022GradSkipCL} introduced GradSkip, which achieves comparable guarantees in the deterministic setting but in terms of local condition numbers, yielding $\smash{\tilde{\mathcal{O}}(\sqrt{\kappa_{\text{max}}})}$ where $\kappa_{\text{max}}$ is the largest condition number among all clients. 
They obtain similar results for the stochastic setting using variance reduction techniques, although a standard stochastic variant is not provided. 
Several subsequent works have combined Scaffnew with additional techniques to further improve communication efficiency, such as partial participation, compression, and quantization \citep{condat2023tamuna,condat2024locodl,yi2024fedcomloc}.

\Cref{table:existing-methods} summarizes FL algorithms that achieve provable improvements using \bbMLS{} without data heterogeneity-bounding assumptions.
To our knowledge, no \bbMLS{} method has been shown to~achieve a provable reduction in the number of communication rounds in the non-smooth convex setting.

\begin{table*}[!th]
    \begin{center}
    \resizebox{\linewidth}{!}{%
        \begin{tabular}{lccccc}
            \toprule
             Algorithm
                & Function Class
                & Smoothness
                & Oracle
                & Iterations  
                & Communication\\
            
            \midrule
            Scaffold \citep{Karimireddy2019SCAFFOLDSC}
                & convex
                & smooth
                & stochastic
                & $\tilde{\mathcal{O}}(\epsilon^{-2})$  
                & $\tilde{\mathcal{O}}(\epsilon^{-1})$ \\
            
            Scaffold \citep{Karimireddy2019SCAFFOLDSC}
                & strongly convex
                & smooth
                & stochastic
                & $\tilde{\mathcal{O}}(\epsilon^{-1})$  
                & $\tilde{\mathcal{O}}(\kappa)$ \\
            GradSkip \citep{Maranjyan2022GradSkipCL}
                & strongly convex
                & smooth
                & deterministic
                & $\tilde{\mathcal{O}}(\kappa_{\text{max}})$  
                & $\tilde{\mathcal{O}}(\sqrt{\kappa_{\text{max}}})$ \\
            
            Scaffnew \citep{Mishchenko2022ProxSkipYL}
                & strongly convex
                & smooth
                & deterministic
                & $\tilde{\mathcal{O}}(\kappa)$  
                & $\tilde{\mathcal{O}}(\sqrt{\kappa})$ \\
            
            Scaffnew \citep{Mishchenko2022ProxSkipYL}
                & strongly convex
                & smooth
                & stochastic
                & $\tilde{\mathcal{O}}(\epsilon^{-1})$  
                & $\tilde{\mathcal{O}}(\epsilon^{-1/2})$ \\
                
            Scaffnew \citep{Hu2023TighterAF}
                & strongly convex
                & smooth
                & stochastic
                & $\tilde{\mathcal{O}}(\epsilon^{-1})$  
                & $\tilde{\mathcal{O}}(\sqrt{\kappa})$ \\
                
            FedMLS (this work)
                & convex
                & non-smooth
                & deterministic
                & $\mathcal{O}(\epsilon^{-2})$  
                & $\mathcal{O}(\epsilon^{-1})$ \\
                
            FedMLS (this work)
                & convex
                & non-smooth
                & stochastic
                & $\mathcal{O}(\epsilon^{-2})$  
                & $\mathcal{O}(\epsilon^{-1})$ \\
                
            \bottomrule \\[-0.85em]
            \multicolumn{6}{l}{\scriptsize$\tilde{\mathcal{O}}$ hides logarithmic factors, $\kappa$ denotes the condition number for the combined objective, $\kappa_{\text{max}}$ is the largest condition number among all clients.}\\
        \end{tabular}
    }
    \end{center}
    \caption{Comparison of algorithms with provable reduction in communication rounds using~\bbMLS{}. 
    }
    \label{table:existing-methods}
\end{table*}

\section{The {\bbFed} Algorithm} \label{sec:mainresult}

At its core, \bbFed applies the Moreau Envelope Projection Efficient Subgradient Method (MOPES) by \citep{Thekumparampil2020ProjectionES} to a product-space reformulation of the FL problem.
We begin with a brief overview of projection-efficient methods.

\subsection{Projection-Efficient Methods}

Consider the following constrained optimization problem: 
\begin{equation} \label{eqn:constrained-problem}
    \min_{X \in \mathcal{X}} ~~ F(X),
\end{equation}
where $F : \mathbb{R}^{d \times n} \to \mathbb{R}$ is a convex Lipschitz continuous function, and $\mathcal{X} \subseteq \mathbb{R}^{d \times n}$ is a convex set. 
Here, we considered a matrix variable to facilitate later connections with FL; otherwise, the approach is not restricted to matrices.

Under standard assumptions, the classical first-order methods like projected (stochastic) subgradient method can find an $\epsilon$-suboptimal solution to this problem in $\mathcal{O}(\epsilon^{-2})$ iterations. 
These methods require computing a (stochastic) subgradient and performing a projection at each iteration. 
However, in many applications, projections can be computationally expensive. 
To address this issue, \citet{Thekumparampil2020ProjectionES} proposed the MOPES algorithm, which leverages a Moreau envelope approximation:
\begin{equation} \label{eqn:constrained-problem-regularized}
    \min_{X \in \mathcal{X}, X' \in \mathcal{X}'} ~ \Psi (X, X') := F(X') + \frac{1}{2\lambda} \norm{X - X'}_F^2,
\end{equation}
where $\mathcal{X}' \subseteq \mathbb{R}^{d \times n}$ is a convex, compact set of choice that admits a simple projection and contains a solution, $X^\star$, to the original problem \eqref{eqn:constrained-problem}, such as a Euclidean norm ball with a sufficiently large radius, $R$. 
By choosing $\lambda > 0$ small, we can control the deviation of the solution of \eqref{eqn:constrained-problem-regularized} from that of~\eqref{eqn:constrained-problem}. 
MOPES applies an accelerated proximal gradient method to solve \eqref{eqn:constrained-problem-regularized}, which performs gradient steps with respect to the term $\smash{\psi(X,X') := \frac{1}{2\lambda} \norm{X - X'}_F^2}$, and proximal steps with respect to the term $\phi(X,X') := F(X') + I_{\sX'}(X') + I_{\sX}(X)$, where $I_{\sX}$ and $I_{\sX'}$ represents indicator functions on $\sX$ and $\sX'$, respectively. 

Since $\phi(X,X')$ is separable in terms of $X$ and $X'$, the prox operator decomposes into two independent sub-problems, one for each variable. 
The prox operator of $I_\mathcal{X}(X)$ corresponds to a Euclidean projection onto $\mathcal{X}$, while the prox operator for $F(X') + I_{\sX'}(X')$ requires solving the sub-problem:
\begin{equation} \label{eqn:prox-subproblem}
    \prox_{F/\beta}(X') = \argmin_{U \in \mathcal{X}'} ~ F(U) + \frac{\beta}{2} \norm{U - X'}_F^2,
\end{equation}
with step-size parameter $\beta > 0$. 
This leads to the following update steps, for $k=1,\ldots,K$:
\begin{equation} \label{eq:mopes-pseudo}
    \left \vert ~~
    \begin{aligned}
        & Y_k = (1-\gamma_k) X_{k-1} + \gamma_k Z_{k-1} 
        \quad \text{and} \quad  Y'_k = (1-\gamma_k) X_{k-1}' + \gamma_k Z_{k-1}' \\
        & Z_k = \proj_{\mathcal{X}}\big(Z_{k-1} - \tfrac{1}{\beta_k \lambda} (Y_k - Y_k') \big) \\
        & (Z_k', \tilde{Z}_k') = \text{approx-}\prox_{F/\beta_k} \big(Z_{k-1}' - \tfrac{1}{\beta_k \lambda} (Y_k' - Y_k) \big) \\
        & X_k = (1-\gamma_k) X_{k-1} + \gamma_k Z_k 
        \quad \text{and} \quad  X_k' = (1-\gamma_k) X_{k-1}' + \gamma_k \tilde{Z}_k'
    \end{aligned}
    \right.
\end{equation}
where the parameters are $\beta_k = \tfrac{4}{\lambda k}$ and $\gamma_k = \tfrac{2}{k+1}$. 

The procedure $\text{approx-}\prox$ produces two outputs, $Z_k'$ and $\tilde{Z}_k'$, representing the final iterate and the weighted average of iterates, respectively, using the following stochastic subgradient method to solve the sub-problem in \eqref{eqn:prox-subproblem}: 
Starting from $U_{k,0} = \tilde{U}_{k,0}= Z_{k-1}'$, and denoting the input by
${V_k := Z_{k-1}' - \tfrac{1}{\beta_k \lambda} (Y_k' - Y_k)},$
the updates proceed as follows for $t = 1, \ldots, T_k$:
\begin{equation} \label{eq:approx-subroutine}
    \left \vert ~~
    \begin{aligned}
        & \hat{U}_{k,t} = U_{k,t-1} - \tfrac{1}{(1+t/2)} \big(\tfrac{1}{\beta_k} \widetilde{\nabla} f (U_{k,t-1}) + U_{k,t-1} - V_k\big) \\
        & U_{k,t} = \proj_{\mathcal{X}'} ( \hat{U}_{k,t} ) \\
        & \tilde{U}_{k,t} = (1-\theta_t) \tilde{U}_{k,t-1} + \theta_t U_{k,t} \quad \text{where} \quad \smash{\theta_t = \tfrac{2(t+1)}{t(t+3)}}.
    \end{aligned}
    \right.
\end{equation}
Here $\smash{\widetilde{\nabla} f (U)}$ represents an unbiased stochastic subgradient estimator with bounded variance, $\sigma^2$.

The following theorem establishes the convergence guarantees of MOPES. 
\begin{theorem}[Theorem 5 in \citet{Thekumparampil2020ProjectionES}]
    \label{thm:conv-guarantee}
    Let $\mathcal{X}'$ be a convex, compact set enclosing a solution to problem \eqref{eqn:constrained-problem}, and suppose $F$ is convex and $G$-Lipschitz continuous on $\mathcal{X}'$. 
    Then, after $K$ iterations of the MOPES algorithm, using a $T_k \geq (4G^2+\sigma^2)\lambda^2K k^2/2D $ iterations of the approx-prox sub-solver with an unbiased stochastic subgradient oracle of bounded variance $\sigma^2$, an estimate $X_K \in \mathcal{X}$ is generated, satisfying
    \begin{align*}
        \bbE\big[ F(X_K) \big] - F(X^\star)
            \leq &\; \frac{10\|X_0-X^\star\|_F^2 +8D}
                      {\lambda K(K+1)} + G^2\frac{\lambda}{2} 
    \end{align*}
    for any choice of $\lambda>0$ and $D>0$.
\end{theorem}

The MOPES procedure allows for a different number of projections onto $\mathcal{X}$ (a total of $K$ times) and first-order oracle calls $\smash{\widetilde{\nabla}f}$ (a total of $\sum_{k=1}^K T_k$ times). 
By choosing $D = \Theta(1)$ and $\lambda = \Theta(\frac{1}{K})$, we obtain $\bbE\big[ F(X_K) \big] - F(X^\star) \leq \mathcal{O}(\frac{1}{K})$. This choice also implies that we can select a $T_k = \Theta(k)$, which gives us $\sum_{k=1}^K T_k = \Theta(K^2)$. 

\subsection{Federated Learning with Multiple Local Steps}
The standard FL problem is typically formulated as:
\begin{align} \label{eqn:FL-model-problem}
    \min_{x \in \mathbb{R}^{d}}
        \; f(x) := \frac{1}{n}\sum_{i=1}^n f_i(x),
\end{align}
where the functions $f_i:\mathbb{R}^d \to \mathbb{R}$, for $i=1,\ldots,n$, are the local loss functions making use of each client’s own dataset and $x \in \mathbb{R}^d$ represents the model parameters to be found. 
We denote by $f:\mathbb{R}^d \to \mathbb{R}$ the joint loss function that the clients aim to minimize collectively. 

Let $x_i \in \mathbb{R}^d$ denote the local weights for client $i$, for $i=1,\ldots,n$. 
To establish the connection between the projection operator and the aggregation step, we first rewrite \eqref{eqn:FL-model-problem} in terms of these local variables. 
For notational convenience, we introduce the matrix $X = [x^1,\ldots, x^n] = \mathbb{R}^{d \times n}$, where the $i$th column of $X$ corresponds to the parameters of client $i$.

Then, we reformulate \eqref{eqn:FL-model-problem}, similar to \citep{Mishchenko2022ProxSkipYL}, as
\begin{equation}
\label{eqn:FL_model_problem_consensus}
    \min_{X \in \mathbb{R}^{d \times n}} ~
   F(X) := \frac{1}{n}\sum_{i=1}^n f_i(x^i)  \quad \mathrm{subject~to} \quad 
    x^1 = \dots = x^n. 
\end{equation}
Problems \eqref{eqn:FL-model-problem} and \eqref{eqn:FL_model_problem_consensus} are equivalent, where the former offers a more compact presentation and the latter explicitly considers the clients' local weights.
Importantly, problem \eqref{eqn:FL_model_problem_consensus} is an instance of \eqref{eqn:constrained-problem} where the constraint is the consensus set
$    \sX = \big\{ X = [x^1, \ldots, x^n] \in \mathbb{R}^{d \times n} \mid x^1 = \cdots = x^n \big\}. 
$

\newpage  

\begin{wrapfigure}[29]{R}{0.5325\textwidth}
    \vspace{-1.0em}
    \begin{minipage}{0.5325\textwidth}
        \begin{algorithm}[H]  
            \caption{\bbT (\bbFed)}
            \label{alg:FedMLS-Alg}
            \begin{algorithmic}[1]    
                \STATE {\bfseries Input:} $R, K, ~ \{\lambda_k\}_{k=1}^K, ~ \{T_k\}_{k=1}^K$.  
                \STATE {\bfseries Parameters:} $\beta_k = \frac{4}{\lambda_k k}$, $\gamma_k = \frac{2}{k+1}$, $\smash{\theta_t = \tfrac{2(t+1)}{t(t+3)}}$ 
                \STATE Set $y_0 \in \mathbb{R}^d$, let 
                        $z_1 = x^i_0=x_0=z_0^i = y_0^i = y_0$
                \FOR{$k=1$ {\bfseries to} $K$}
                    \STATE Server sends $y_k$ to the clients

                    \vspace{0.25em}
                    --- Clients $i=1, \dots, n$ start local training ---
                    \vspace{0.25em}
                    \STATE $v^i_k = z^i_{k-1} - \tfrac{1}{\beta_k \lambda_k} (y^i_k - y_k)$ 
                    \STATE $u^i_{k,0}= \tilde{u}^i_{k,0} =z^i_{k-1} $
                    \FOR{$t=1$ {\bfseries to} $T_k$}
                        \STATE $\hat{u}^i_{k,t} = u^i_{k,t-1} - \tfrac{1}{(1+t/2)} \big(\tfrac{1}{n\beta_k} \widetilde{\nabla} f_i (u^i_{k,t-1})$
                        \vspace{-0.1em}      \qquad\qquad\qquad\qquad\qquad\quad\; $+\, u^i_{k,t-1} - v^i_k\big)$
                         \STATE $u^i_{k,t} = \hat{u}^i_{k,t} \cdot \min\hspace{-0.15em}\big(1, \tfrac{R}{\|\hat{u}^i_{k,t}\|}\big)$
                        \STATE $\tilde{u}^i_{k,t} = (1-\theta_t) \tilde{u}^i_{k,t-1} + \theta_t u^i_{k,t}$ 
                    \ENDFOR
                    \vspace{0.1em}
                    \STATE $z^i_{k} = u^i_{k,T_k}$ and $\tilde{z}^i_{k} = \tilde{u}^i_{k,T_k}$
                    \STATE $x^i_k = (1-\gamma_k) x^i_{k-1} + \gamma_k \tilde{z}^i_{k}$
                    \STATE $y^i_{k+1} = (1-\gamma_{k+1}) x^i_{k} + \gamma_{k+1} z^i_{k}$
                    \STATE Clients send $y^i_{k+1}$ to the server

                    \vspace{0.25em}
                    --- End of local training phase ---
                    \vspace{0.25em}
                    \STATE $x_k = (1-\gamma_k) x_{k-1} + \gamma_k z_{k}$
                     \STATE $y_{k+1} = (1-\gamma_{k+1}) x_{k} + \gamma_{k+1} z_{k}$
                    \STATE $\textstyle{z_{k+1} = z_{k} \hspace{-0.1em}-\hspace{-0.1em} \frac{1}{\beta_{k+1}\lambda_{k+1}} ( y_{k+1} - \frac{1}{n} \sum_{i=1}^n y^i_{k+1} )}$
                \ENDFOR
                \STATE \textbf{return} $x_K$
            \end{algorithmic}
        \end{algorithm}
    \end{minipage}
\end{wrapfigure}

Moreover, we select 
$    \sX' = \big\{ X = [x^1, \ldots, x^n] \in \mathbb{R}^{d \times n} \mid \|x^i\| \leq R, ~\text{for} ~ i=1,\dots,n \big\}. 
$
This ensures $\sX'$ is separable across columns of $X$, which becomes important in the analysis later.

We now apply MOPES, as described in \eqref{eq:mopes-pseudo}, to solve \eqref{eqn:FL_model_problem_consensus}.
The projection in the second step can be computed by
\begin{equation} \label{eq:proj-agg}
    \proj_{\sX}(X)
        = \bigg(\frac{1}{n} \sum_{i=1}^n x^i \bigg) \mathbb{1}_{n}^{\top},
\end{equation}
where $\mathbb{1}_{n}$ denotes a vector of ones.
This projection corresponds to the client-server communication.
The columns of $Z_k$ (in \eqref{eq:mopes-pseudo}) are identical by definition of this projection, and it is easy to verify that the columns in $Y_k$ and $X_k$ are also identical. 
We represent them using vectors, $z_k, y_k, x_k \in \mathbb{R}^d$, such that $Z_k = z_k \mathbb{1}_n^\top$, $Y_k = y_k \mathbb{1}_n^\top$, and $X_k = x_k \mathbb{1}_n^\top$.
In contrast, in the clients, the columns of $X'_k, Y'_k, Z'_k$ and $\smash{\tilde{Z}'_k}$ are distinct.

For brevity in what follows, we will omit the prime notation ($'$) when referring to columns, since the client index distinguishes between variables corresponding to $X_k$ and those to $X'_k$.

We now rewrite the algorithm using these column expressions.
The interpolation steps on the first and last lines of MOPES in \eqref{eq:mopes-pseudo} are straightforward:
\begin{equation*}
    \begin{aligned}
        y_k = (1-\gamma_k) x_{k-1} + \gamma_k z_{k-1} 
        ~~~&\text{and}~~~ y^i_k = (1-\gamma_k) x^i_{k-1} + \gamma_k z^i_{k-1}, ~~~ \text{for $i = 1,\dots,n$};\\
        x_k = (1-\gamma_k) x_{k-1} + \gamma_k z_{k} 
        ~~~&\text{and}~~~ x^i_k = (1-\gamma_k) x^i_{k-1} + \gamma_k \tilde{z}^i_{k}, ~~~ \text{for $i = 1,\dots,n$}.
    \end{aligned}
\end{equation*}
We also derive the projection step as
\begin{equation*}
    z_k = \frac{1}{n} \sum_{i=1}^n \Big(z_{k-1} - \frac{1}{\beta_k\lambda} (y_k - y^i_k) \Big) = z_{k-1} - \frac{1}{\beta_k\lambda} \Big( y_k - \frac{1}{n} \sum_{i=1}^n y^i_k \Big).
\end{equation*}
Since both the objective function $F$ and the constraint $\sX'$ are columnwise separable, the prox sub-problem can be decomposed and solved independently for each column (\textit{i.e.}, each client) in~parallel:
\begin{equation*}
    (z^i_k, \tilde{z}^i_k) = \text{approx-}\prox_{\frac{1}{n} f_i/\beta_k} \big(z^i_{k-1} - \tfrac{1}{\beta_k \lambda} (y^i_k - y_k)\big).
\end{equation*}
Denoting $v^i_k := z^i_{k-1} - \tfrac{1}{\beta_k \lambda} (y^i_k - y_k)$, the subroutine in \eqref{eq:approx-subroutine} becomes
\begin{equation} 
    \left \vert ~~
    \begin{aligned}
        & \hat{u}^i_{k,t} = u^i_{k,t-1} - \tfrac{1}{(1+t/2)} \big(\tfrac{1}{n\beta_k} \widetilde{\nabla} f_i (u^i_{k,t-1}) + u^i_{k,t-1} - v^i_k\big) \\
        & u^i_{k,t} = \hat{u}^i_{k,t} \cdot \min\hspace{-0.15em}\big(1, \tfrac{R}{\|\hat{u}^i_{k,t}\|}\big) \\
        & \tilde{u}^i_{k,t} = (1-\theta_t) \tilde{u}^i_{k,t-1} + \theta_t u^i_{k,t} \quad \text{where} \quad \smash{\theta_t = \tfrac{2(t+1)}{t(t+3)}}.
    \end{aligned}
    \right.
\end{equation}
Here, the second line projects $\hat{u}^i_{k,t}$ onto the Euclidean norm ball of radius $R$ centered at the origin. 
This parallelized implementation of approx-prox subroutine corresponds to the local training steps. 

As a final adjustment, we can align the algorithm's flow with the general convention in FL by starting with the approx-prox step and ending with the projection step---that is, beginning with local updates and concluding with aggregation. 
This swap is possible since the first three steps at $k=1$ are redundant under the initialization $z_0 = z^1_0 = \dots = z^n_0$ and $\gamma_1 = 1$, which gives:
\begin{equation*}
\begin{aligned}
    & y_1 = (1-\gamma_1) x_{0} + \gamma_1 z_{0} = z_0 \\
    & y^i_1 = (1-\gamma_1) x^i_{0} + \gamma_1 z^i_{0} = z^i_0 = z_0 ~~~ \text{for $i = 1,\dots,n$} \\
    & \textstyle{z_1 = z_0 - \frac{1}{\beta_{1}\lambda} ( y_{1} - \frac{1}{n} \sum_{i=1}^n y^i_{1} ) = z_0} ~~~ \text{for $i = 1,\dots,n$} 
\end{aligned}
\end{equation*}
Hence, starting from the initialization $z_1 = z_0 = z_0^1 = \dots = z_0^n$, the algorithm proceeds as follows:
\begin{equation*}
\left \vert ~~
    \begin{aligned}
        & (z^i_k, \tilde{z}^i_k) = \text{approx-}\prox_{\frac{1}{n} f_i/\beta_k} (z^i_{k-1} - \tfrac{1}{\beta_k \lambda} (y^i_k - y_k)) ~~~ \text{for $i = 1,\dots,n$} \\
        & x_k = (1-\gamma_k) x_{k-1} + \gamma_k z_{k} 
        ~~~\text{and}~~~ x^i_k = (1-\gamma_k) x^i_{k-1} + \gamma_k \tilde{z}^i_{k} ~~~ \text{for $i = 1,\dots,n$}\\
        & y_{k+1} = (1-\gamma_{k+1}) x_{k} + \gamma_{k+1} z_{k} 
        ~~~\text{and}~~~ y^i_{k+1} = (1-\gamma_{k+1}) x^i_{k} + \gamma_{k+1} z^i_{k} ~~~ \text{for $i = 1,\dots,n$} \\
        & \textstyle{z_{k+1} = z_{k} - \frac{1}{\beta_{k+1}\lambda} ( y_{k+1} - \frac{1}{n} \sum_{i=1}^n y^i_{k+1} ).}
    \end{aligned}
\right.
\end{equation*}
The final design of \bbFed{} is presented in \Cref{alg:FedMLS-Alg}. The following result, derived from \Cref{thm:conv-guarantee}, establishes its complexity guarantees. 

\begin{corollary} \label{cor:convg-guar}
    Let the client loss functions, $f_i$, be $G$-Lipschitz continuous convex functions, equipped with an unbiased stochastic subgradient oracle with bounded variance, $\sigma^2$. 
    Fix a target $\epsilon > 0$ and choose $\lambda_k = \lambda = \epsilon/G^2$ and $T_k = T = \lceil 6^2 \sqrt{3n} \norm{x_0 - x^*} (4G^2 + \sigma^2)/(G \epsilon)  \rceil$, both fixed. 
    Then, after at mosth $K = \lceil 6 \sqrt{n} \norm{x_0 - x^*}G / \epsilon \rceil$ communication rounds, FedMLS finds an estimate satisfying 
       $\bbE[ f(x_K) ] - f(x^*) \leq \epsilon.$ 
    In total, this results in $KT = \Theta(\epsilon^{-2})$ local steps per client. 
\end{corollary}

\begin{remark} \label{rem:parameters}
    Fixing $\epsilon$ a priori is often impractical, and constants such as $\norm{x_0 - x^*}$ are usually unknown.
    Theoretical insights suggest choosing $\lambda = \Theta(1/K)$ and $T = \Theta(K)$. 
    Instead of using fixed values, we adopted a practical approach and set $\lambda_k = \lambda_0 / k$ and $T_k = T_0 k$ for some $\lambda_0, T_0 > 0$. 
\end{remark}

\section{Numerical Experiments} \label{sec:numerical_exp}

The performance of the proposed algorithm was evaluated using the Wisconsin Breast Cancer dataset \citep{wolberg1990multisurface}, consisting of 699 samples and 10 features. Missing values were imputed using column-wise means.
To simulate a heterogenous FL setting, we partitioned data among $n=10$ clients using k-means clustering.
A Support Vector Machine (SVM) was trained by solving 
\begin{equation} \label{eqn:svm-formulation}
    \min_{x \in \mathbb{R}^d, \, \theta \in \mathbb{R}} ~~ \frac{1}{n} \sum_{i=1}^n \sum_{j=1}^{m_i} \max\hspace{-0.15em}\big(0, 1-b^i_j (x^\top a^i_j + \theta) \big),
\end{equation}
where $\{(a^i_j, b^i_j)\}_{j=1}^{m_i}$ represents the dataset of the {$i^\text{th}$} client, with $a_j^i \in \mathbb{R}^d$ denoting the numerical features and $b_j^i \in \{-1, 1\}$ binary labels that indicate the class of each sample.
To obtain the ground truth solution, we first solved the problem using CVX~\citep{grant2014cvx}.

\begin{figure*}[!t]
    \centering
    \includegraphics[width=\linewidth]{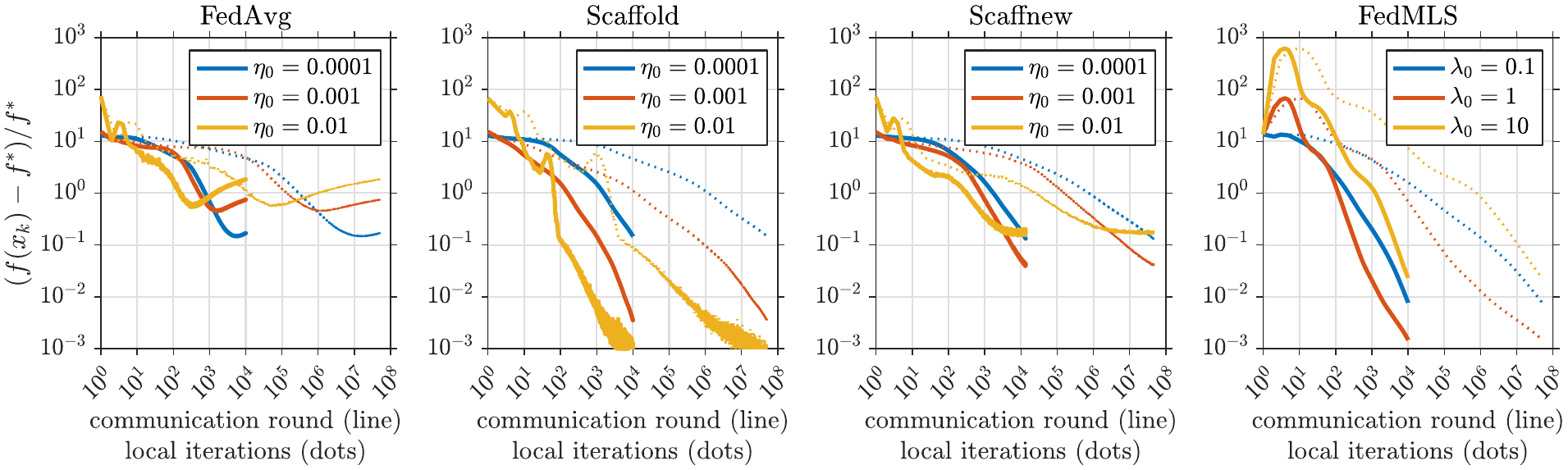}
    \caption{Comparing FedAvg, Scaffold, Scaffnew, and FedMLS on the SVM problem. 
    FedMLS avoids client drift and converges smoothly to the ground truth with theoretical guarantees.}
    \label{fig:svm}
\end{figure*}

We applied FedMLS, Scaffold, Scaffnew, and FedAvg to solve \eqref{eqn:svm-formulation}, initializing all parameters to zero.
The clients used stochastic subgradients, in mini-batches with random 10\% of their samples. The algorithms were run with 20 different random seeds, and the results averaged.
For FedMLS, Scaffold, and FedAvg, we used an increasing number of local steps, with $T_k = k$ at communication round $k$. 
Scaffnew does not have a predefined number of local iterations; instead, it aggregated the models at iteration $t$ with probability $\smash{p_t = 1/\sqrt{t}}$.
The standard parameters for FedAvg, Scaffold, and Scaffnew are ineffective in the non-smooth convex setting. 
To address this, we adjusted their parameters through trial and error, experimenting with both constant and decreasing learning rates, given the problem is non-smooth. 
Whether these methods can be theoretically extended to the non-smooth setting, and whether our parameter choices are optimal, remains an open question beyond the scope of this paper. 
Specifically, for FedAvg, we let $\smash{\eta_k = \eta_0 / \sqrt{k}}$, the standard choice for the subgradient method to ensure convergence in convex optimization.
FedAvg did not converge to a solution, exhibiting a client drift, as seen in \Cref{fig:svm}.
For Scaffold, we followed \citet{Karimireddy2019SCAFFOLDSC} and used global $\eta_g = \sqrt{n}$ and local $\eta_t = \eta_0 / (\eta_g t)$, for $t$ total local steps; see \Cref{fig:svm}.
For Scaffnew, we let $\eta_t = \eta_0 / \sqrt{t}$; see \Cref{fig:svm}.
For FedMLS, we used $\lambda_k = \lambda_0 / k$, as motivated in \Cref{rem:parameters}.
The hyper-parameters, $\eta_0$ and $\lambda_0$, were tuned in powers of $10$.

\section{Conclusions and Future Directions}

We presented FedMLS, which achieves a provable improvement in communication rounds without requiring data heterogeneity bounding assumptions in the general convex non-smooth setting; analogous to those of Scaffold \citep{Karimireddy2019SCAFFOLDSC} and Scaffnew \citep{Mishchenko2022ProxSkipYL} for the smooth convex (stochastic) and strongly convex settings (stochastic and deterministic), respectively. 
While the existing theory for \bbFed{} focuses on the non-smooth setting, our preliminary analysis suggests that \bbFed{} achieves an $\mathcal{O}(1/\sqrt{\epsilon})$ communication complexity in the convex and smooth deterministic setting, marking a novel improvement.
Additionally, a natural extension is to analyze partial client participation, which corresponds to a block-coordinate variant of the underlying optimization technique. 
Finally, a key direction for future work is to conduct comprehensive experiments to further validate the practical performance of FedMLS across diverse FL settings.

\section*{Acknowledgments}

This work was supported by the Wallenberg AI, Autonomous Systems and Software Program (WASP) funded by the Knut and Alice Wallenberg Foundation. 
Alp Yurtsever and Ali Dadras further acknowledge support from the Swedish Research Council, under registration number 2023-05476. 
We thank the High Performance Computing Center North (HPC2N) at Umeå University for providing computational resources and valuable support during test and performance runs. 
The computations were enabled by resources provided by the National Academic Infrastructure for Supercomputing in Sweden (NAISS), partially funded by the Swedish Research Council through grant agreement no. 2022-06725.

\bibliography{arxiv}
\bibliographystyle{iclr2025_conference}

\end{document}